%% file: main.tex
\documentclass[letterpaper]{article} % DO NOT CHANGE THIS
\usepackage{aaai23}  % DO NOT CHANGE THIS
\usepackage{times}  % DO NOT CHANGE THIS
\usepackage{helvet}  % DO NOT CHANGE THIS
\usepackage{courier}  % DO NOT CHANGE THIS
\usepackage[hyphens]{url}  % DO NOT CHANGE THIS
\usepackage{graphicx} % DO NOT CHANGE THIS
\usepackage{natbib}  % DO NOT CHANGE THIS AND DO NOT ADD ANY OPTIONS TO IT
\usepackage{caption} % DO NOT CHANGE THIS AND DO NOT ADD ANY OPTIONS TO IT
\usepackage{algorithm}
\usepackage{algorithmic}
\usepackage{xspace}
\usepackage{caption}
\usepackage{subcaption}
\usepackage{booktabs}
\newcommand{\name}{TS-TrajGen\xspace}
\usepackage{amsthm}
\usepackage{amsmath}
\usepackage{amssymb}
\usepackage{multirow}
\usepackage{bm}
\newtheorem{definition}{Definition}

\usepackage{newfloat}
\usepackage{listings}
\DeclareCaptionStyle{ruled}{labelfont=normalfont,labelsep=colon,strut=off} % DO NOT CHANGE THIS
\floatstyle{ruled}
\newfloat{listing}{tb}{lst}{}
\floatname{listing}{Listing}
\title{Continuous Trajectory Generation Based on Two-Stage GAN}
\author{
    Wenjun Jiang\textsuperscript{\rm 1},
    Wayne Xin Zhao\textsuperscript{\rm 4},
    Jingyuan Wang\textsuperscript{\rm 1, \rm 2, \rm 3}\thanks{Corresponding author: jywang@buaa.edu.cn\\},
    Jiawei Jiang\textsuperscript{\rm 1}
}
\affiliations{
    \textsuperscript{\rm 1} School of Computer Science and Engineering, Beihang University, Beijing, China\\
    \textsuperscript{\rm 2} Pengcheng Laboratory, Shenzhen, China\\
    \textsuperscript{\rm 3} School of Economics and Management, Beihang University, Beijing, China\\
    \textsuperscript{\rm 4} Gaoling School of Artificial Intelligence, Renmin University of China, Beijing, China
}

\usepackage{bibentry}
\begin{document}

\maketitle

\begin{abstract}
Simulating the human mobility and generating large-scale trajectories are of great use in many real-world applications, such as urban planning, epidemic spreading analysis, and geographic privacy protect. Although many previous works have studied the problem of trajectory generation, the continuity of the generated trajectories has been neglected, which makes these methods useless for practical urban simulation scenarios. To solve this problem, we propose a novel two-stage generative adversarial framework to generate the continuous trajectory on the road network, namely \name, which efficiently integrates prior domain knowledge of human mobility with model-free learning paradigm. Specifically, we build the generator under the human mobility hypothesis of the A* algorithm to learn the human mobility behavior. For the discriminator, we combine the sequential reward with the mobility yaw reward to enhance the effectiveness of the generator. Finally, we propose a novel two-stage generation process to overcome the weak point of the existing stochastic generation process. Extensive experiments on two real-world datasets and two case studies demonstrate that our framework yields significant improvements over the state-of-the-art methods.
\end{abstract}

\input{introduction}
\input{problem_statement}
\input{the_proposed_model}

\input{experiment}
\input{related_work}
\input{conclusion}

\section{Acknowledgments}
This work was supported by the National Key R\&D Program of China (2021ZD0111201). Prof. Wang’s work was supported by the National Natural Science Foundation of China (No. 82161148011, 72222022, 72171013), the Fundamental Research Funds for the Central Universities (YWF-22-L-838) and the DiDi Gaia Collaborative Research Funds. Prof. Zhao’s work was supported by the National Natural Science Foundation of China (No. 62222215).

\bibliography{main}

\end{document}

%% file: introduction.tex
\section{Introduction}

%background to topic
Modeling human mobility and generating synthetic yet realistic trajectories is crucial in many applications~\cite{a_survey_on_deep_learninng}. On the one hand, synthetic trajectories are fundamental for urban planning, epidemic spreading analysis, and traffic control, e.g., simulating changes in urban mobility in the presence of new infrastructure or international events~\cite{8958668, libcity}. On the other hand, generating synthetic trajectories is a viable solution to protect the privacy of human mobility trajectory data~\cite{10.1145/3106774}. The difficulty of open source sharing of trajectory data today is heavily due to privacy concerns, which hinders most existing data-driven studies. Using synthetic trajectories instead of real trajectories can not only preserve the utility of trajectory data, but also avoid leakage of user privacy. Therefore, in such cases, it is important to generate synthetic trajectories with good data utility.

In the last decades, the problem of trajectory generation has been widely studied. In the early stage, the researchers aim to build model-based methods to model the regularity of human mobility~\cite{epr, TimeGeo}, such as temporal periodicity, spatial continuity. These methods assume that human mobility can be described by specific mobility patterns and thus can be modeled with finite parameters with explicit physical meanings. However, in fact, human mobility behaviors exhibit complex sequential transition regularities, which can be time-dependent and high-ordered. Thus, although these model-based methods have the advantage of being interpretable by design, their performance is limited due to the simplicity of the implemented mechanisms~\cite{a_survey_on_deep_learninng}. To address the above limitation, the researchers propose model-free methods~\cite{10.1155/2018/9203076, location_distribution}, mainly using neural network generative paradigms such as generative adversarial network and variational autoencoder. Unlike model-based methods, model-free methods abandon the extraction of specific human mobility patterns, and instead directly build a neural network to learn the distribution of the real data and generate trajectories from the same distribution.

% Despite the inspiring results of model-free methods, there are several key challenges remaining to be solved: (1) First, to our best knowledge, existing methods are unable to directly generate fine-grained trajectories based on road networks, which requires ensuring the continuity and integrity of trajectories. All the methods mentioned above generate either discontinuous POI trajectories or discontinuous GPS trajectories, which makes these synthetic trajectories unusable for downstream applications such as traffic simulation, urban planning, and epidemic spreading analysis. (2) Second, existing methods only focus on modeling individual mobility behaviors, while ignoring global urban traffic constraints. Therefore, the synthetic trajectory data is difficult to restore the real urban traffic conditions like traffic OD flow, which leads to global distortion of the simulation results. 

However, comparing the above approaches, there are several key challenges remaining to be solved: (1) First, the problem of the continuity of the generated trajectories is ignored. The trajectories generated by current methods are not continuous routes on the road network, which makes these synthetic trajectories unusable for downstream applications like traffic simulation. (2) Second, those model-free methods without utilizing prior knowledge of human mobility fail to effectively generate continuous trajectories. (3) Third, the stochastic generation process of existing methods has the error accumulation problem, where the trajectory is randomly generated according to the probability given by the generator. However, once the generator predicts wrong, the process continues to generate under the wrong premise, which reduces the quality of the generated trajectory.
In this paper, we propose \name, a novel two-stage generative adversarial framework with spatial-temporal oriented designs in the generator, discriminator and generation process to tackle the above challenges. Specifically, we build the generator based on the human mobility hypothesis of the A* algorithm~\cite{A_star} to address the first challenges. The A* algorithm is a heuristic search algorithm used extensively on the road network. In the A* hypothesis, human mobility behaviors are determined by two factors: the observed cost from the source road to the candidate road, and the expected cost from the candidate road to the destination. Combining above two costs, A* algorithm evaluates which candidate road is the best candidate to be searched, and then generates a heuristic best continuous trajectory. Therefore, our generator consists of two parts: an attention-based network to learn the observed cost, and a GAT-based network to estimate the expected cost. As for the second challenge, we combine the sequential reward with the mobility yaw reward to improve the effectiveness of the generator, which distinguish the trajectory from the perspective of time series similarity and spatial similarity respectively. For the third challenge, we propose a two-stage generation process to solve the problem of error accumulation. In the first stage, we construct the structural regions on the top of the road network and then generate the regional trajectory. In the second stage, we generate the continuous trajectory in the guidance of the regional trajectory.

To the best of our knowledge, we are the first to solve the continuous trajectory generation problem on the urban road network, through combining the A* algorithm with neural network. In addition, to improve the effectiveness and efficiency of the generation, we build a discriminator combining the sequential reward with mobility yaw reward and propose a two-stage generation process. Extensive experiments on two real-world trajectory data have demonstrated the effectiveness and robustness of our proposed framework.

% To the best of our knowledge, we are the first to solve the fine-grained trajectory generation problem on the urban road network, which means that our framework can be directly applied to practical downstream applications such as urban traffic simulation, urban planning, etc. 
%Experiment Overview

%Roadmap

%% file: problem_statement.tex
\section{Preliminary}

% Human mobility data can be defined as a spatial-temporal trajectory $T = \{x_1, x_2, \ldots, x_n\}$, where the $i_{th}$ element $x_i$ is a spatial-temporal point defined as a tuple $(l_i, t_i)$. The $l_i$ denotes the location identification, which corresponds to a road segment in the road network. And the $t_i$ is the timestamp of $x_i$. Based on the above notation, the fine-grained mobility trajectory can be defined as follow:
% The continuous mobility trajectory can be defined as a time-ordered sequence $T = \{x_1, x_2, \ldots, x_n\}$, where each pair $(x_i, x_{i+1})$ in $T$ should be adjacent in the road network. The $i_{th}$ element $x_i$ is a spatial-temporal point defined as a tuple $(l_i, t_i)$, where the $l_i$ is the road segment ID and the $t_i$ is the timestamp of $x_i$. Based on the above notations, the problem of the continuous mobility trajectory generation can be defined as follows:
A trajectory in the road network can be defined as a time-ordered sequence $T = \{x_1, x_2, \ldots, x_n\}$, where $x_i$ is a spatial-temporal point defined as a tuple $(l_i, t_i)$. The $l_i$ is the road segment ID and the $t_i$ is the timestamp of $x_i$. The continuous trajectory is a trajectory, where each segment pair $(x_i, x_{i+1})$ is adjacent in the road network graph.
% The continuous mobility trajectory can be defined as a time-ordered sequence $T = \{x_1, x_2, \ldots, x_n\}$, which describes the continuous movements of a person from a certain origin road segment to a destination road segment on the road network over a period of time. The $i_{th}$ element $x_i$ is a spatial-temporal point defined as a tuple $(l_i, t_i)$, where the $l_i$ is the road segment ID and the $t_i$ is the timestamp of $x_i$. To describe the continuity, each pair $(x_i, x_{i+1})$ in $T$ should be adjacent in the road network. Based on the above notation, the problem of the fine-grained mobility trajectory generation can be defined as follow:

% \begin{definition}
% \label{fine_grained_trajectory}
% \textbf{Fine-grained Mobility Trajectory.} 
% \end{definition}

\begin{definition}
\label{continuous_mobility_trajectory_generation}
\textbf{Continuous Mobility Trajectory Generation.} Given a real-world mobility trajectory dataset, generate a continuous mobility trajectory $\hat{T} = \{\hat{x}_1, \hat{x}_2, \ldots, \hat{x}_n\}$ with a $\theta$-parameterized generative model $G$.
% through a Markov decision-making process:
% \begin{equation}
% p_{\theta}(\hat{T})=\prod_{i=1}^{n} p_{\theta}\left(\hat{x}_{i} \mid \hat{x}_{1}, \ldots, \hat{x}_{i-1}\right)
% \end{equation}
% where $p_{\theta}$ denotes the generation distribution from generator $G$. The generation of trajectory $\hat{T}$ via generator $G$ is a sequential decision process.
\end{definition}

In general, the process of continuous mobility trajectory generation can be regarded as a Markov decision process (MDP). The state $s$ describes the current movement situation, composed of the current partial trajectory $x_{1:i}$ and the travel destination $l_d$. The action $a$ is the next candidate road segment $l_{j}$ to move. The agent is the generative model, which model the human movement policy $\pi(a | s)$.
% The reward $R$ describes the expected reward for performing an action $a$ in a state $s$, i.e. an evaluation of how good the policy $\pi(a | s)$ is.

\begin{definition}
\label{movement_policy}
\textbf{Human Movement Policy.} The human movement policy $\pi(a | s)$ is the conditional probability of taking action $a$ in the state $s$, which can be formulated as follows:
\begin{equation}
\small
    \pi(a | s) = P(a|s) = P(l_j | x_{1:i}, l_d).
\end{equation}
\end{definition}

Thus, the generation process can be described as maximizing the probability of the generated trajectory as

\begin{equation}
\small
%   \hat{T} = \max \{ \hat{T}^\prime \mid \hat{T}^\prime = \prod_{i=1}^{n} \pi(a_i | s_i)\}.
\hat{T} = \max \prod_{i=1}^{n} \pi(a_i | s_i) = \max_{P_\theta}  \prod_{i=1}^{n} P_\theta(l_{i+1} | x_{1:i}, l_d).
\end{equation}
% = \max  \{ \hat{T} \mid \hat{T} = \prod_{i=1}^{n} P(l_{i+1} | x_{1:i}, l_d)\}

% Thus, the stochastic generation process can be described as a stochastic state transition process, where the probability of generating trajectory $\hat{T}$ is as follows:

% \begin{equation}
% \small
%     P_{\theta}(\hat{T})=\prod_{i=1}^{n} \pi(a_i | s_i) = \prod_{i=1}^{n} P(l_{i+1} | x_{1:i}, l_d).
% \end{equation}

% , that is, the transition process from an initial state (starting from origin road segment) to a stationary state (arriving at destination road segment).

% \begin{algorithm}[tb]
% \caption{Traditional Stochastic Generation Process}
% \label{traditional_generation_process}
% \textbf{Input}: The initial state\\
% \textbf{Output}: The generated trajectory
% \begin{algorithmic}[1] %[1] enables line numbers
% \STATE Let current state $s= (x_0, l_d)$.
% \WHILE{condition}
% \STATE Do some action.
% \IF {conditional}
% \STATE Perform task A.
% \ELSE
% \STATE Perform task B.
% \ENDIF
% \ENDWHILE
% \STATE \textbf{return} solution
% \end{algorithmic}
% \end{algorithm}

%% file: the_proposed_model.tex
\section{The Proposed Framework}

\begin{figure}[tbp]
\centering
\includegraphics[width=0.95\linewidth]{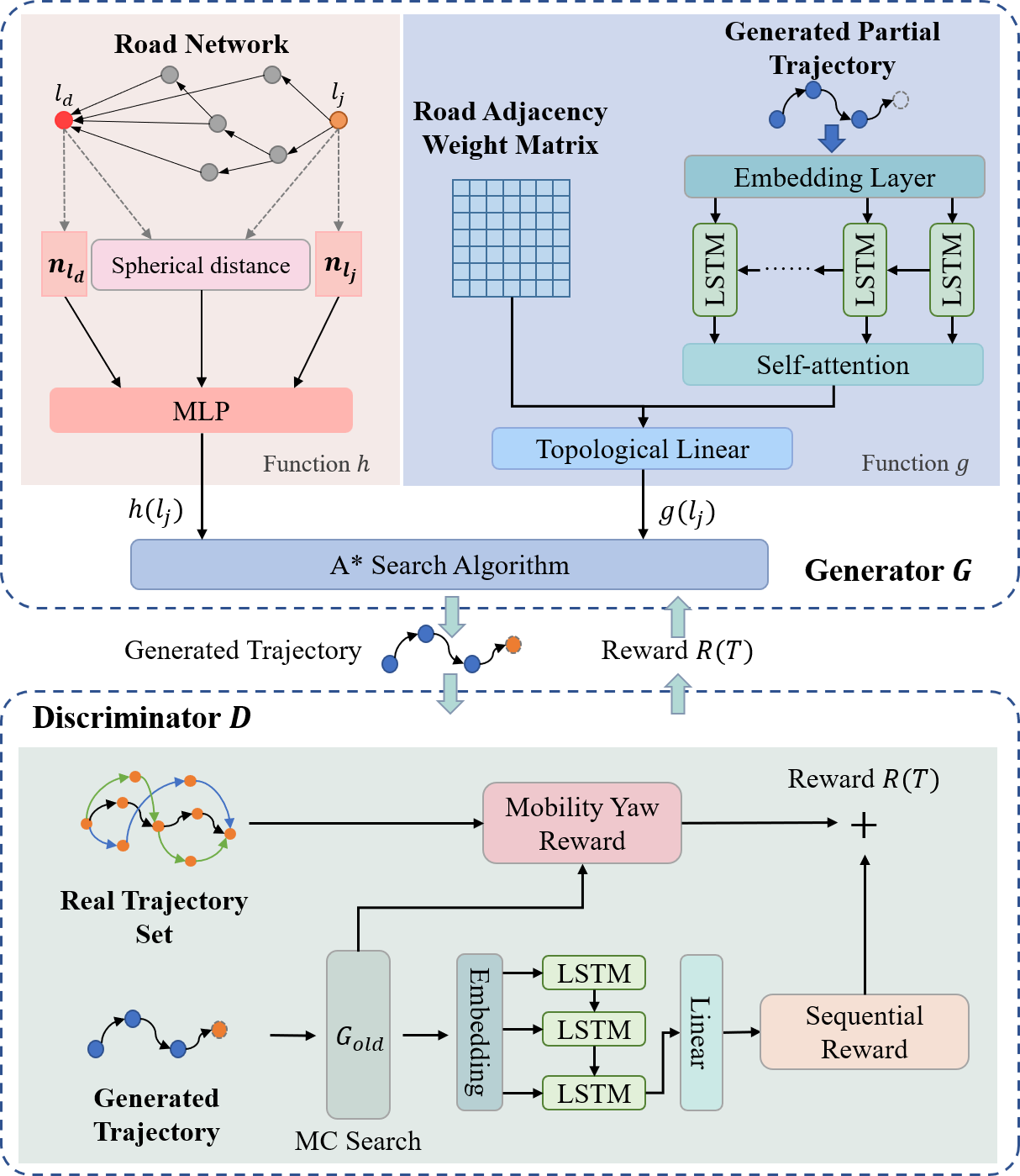}
\caption{The overall generative adversarial framework of \name.}
\label{illustration_of_gan}
\end{figure}

In this section, we first give a brief overview of our proposed framework and then describe each part of our framework in detail. We build a generator $G$ to learn the human movement policy $\pi(a | s)$, and a discriminator $D$ to generate the reward $R$ to guide the optimization process of the generator, which is represented in Figure~\ref{illustration_of_gan}. Besides, we propose a novel two-stage generation process to generate trajectories more accurately and efficiently. Finally, we introduce the training process of our proposed framework.
% instead of using stochastic generation process,

% Since the fine-grained mobility trajectory generation problem can be viewed as a Markov decision process, we leverage the adversarial generative framework to handle the problem. 

% introduce the overview of generator
\subsection{Modeling Human Movement Policy by Generator}

As stated in Definition~\ref{movement_policy}, the key factor that determines the human movement policy $\pi(a | s)$ is the current state $s$, which is described by current partial trajectory $x_{1:i}$ and the travel destination $l_d$. To consider both the impact of the current partial trajectory and the impact of the travel destination, we adopt the hypothesis of A* algorithm to model the human movement policy $\pi(a | s)$. The A* algorithm is a heuristic search algorithm used extensively on the road network, which  evaluates candidate road $l_j$ exactly based on the above two impacts as

%  Thus, to model the mobility policy, we need to consider both the impact of current partial trajectory and the impact of travel destination. In this slight, we adopt the heuristic idea of A* algorithm to model the mobility policy, because A* evaluates candidate road $l_j$ exactly based on the above two impacts as 

\begin{equation}
\label{cost_function_of_a_star}
\small
    f(l_j) = g(l_j) + h(l_j),
\end{equation}

\noindent where the function $g(l_j)$ evaluates the observed cost of the current partial trajectory, and the function $h(l_j)$ estimates the expected cost of arriving the destination $l_d$ through the candidate road $l_j$.
%  from the source to candidate road $l_j$

Combining above two costs, A* algorithm learn the human movement policy as follows:

\begin{equation}
\small
    f(l_j) = - \log \pi(a|s) = - \log P(l_j | x_{1:i}, l_d).
\end{equation}

With the function $f$, A* algorithm evaluates which candidate road is the best candidate to be searched, and then generates the heuristic best continuous trajectory.

However, there are two weak points for generating trajectories with the original A* algorithm. First, the original functions $g$ and $h$ are calculated based on the spherical distance between road segments, which makes it difficult to learn diverse human movement policy. Second, the spherical distance can not accurately estimate the expected cost. For example, although roads $l_i$ and $l_j$ are close in space, they may be one-way roads in two different directions on the road network, which makes the real expected cost between them greater than the spherical distance. Thus, we implement the functions $g$ and $h$ with the neural network to learn the human movement policy.

\subsubsection{Function \textit{g}($\cdot$).} To model the observed cost, we combine the LSTM network with the self-attention mechanism to learn the current moving state and then predict the observed cost with a topologically constrained linear layer.
%  an intuitive idea is to capture the current moving state. In this light, 

First, given a current partial trajectory $x_{1:i}$, we extract the spatial-temporal information through embedding each point $x_i=(l_i, t_i)$ into a dense vector $\bm{x}_i$ as
% shown in Eq.~\ref{traj_embed}.
\begin{equation}
\label{traj_embed}
\small
\bm{x}_i = \operatorname{Embed}(l_i) \mid\mid \operatorname{Embed}(\operatorname{encode}(t_i)),
\end{equation}
\noindent where we encode the timestamp $t_i$ by mapping the weekday timestamp to a group of $24\times60$ time slots (1 to 1440), and the weekend timestamp to another group time slots (1441 to 2880).

% of $24\times60$
After embedding, we transform the current partial trajectory into a sequence of dense vectors $\{\bm{x}_1, \bm{x}_2, \ldots, \bm{x}_i\}$. Then, we apply the LSTM network~\cite{lstm} to model the sequential trajectory moving state $\{\bm{h}_1, \bm{h}_2, \ldots, \bm{h}_i\}$. We further leverage the dot-product attention mechanism~\cite{attention_is_all_you_need} to enhance the latent moving state $\bm{\widetilde{h}}_i$ in the form of
% , since the LSTM may suffer from the long-term dependencies
%Eq.~\ref{enhance_with_attention}.
% As shown in Eq. \ref{enhance_with_attention}, the $\mathbf{\widetilde{h}}_i$ is a weighted sum of moving state sequence $\{\mathbf{h}_1, \mathbf{h}_2, \ldots, \mathbf{h}_i\}$, where the weights are calculated by the function $\operatorname{att}(\cdot,\cdot)$.

% \begin{equation}
% \label{lstm_representation}
% \mathbf{h}_i = \operatorname{LSTM}(\mathbf{x}^e_i, \mathbf{h}_{i-1}),
% \end{equation}

\begin{equation}
\label{enhance_with_attention}
\small
\bm{\widetilde{h}}_i = \sum_{k=1}^i \operatorname{att}(\bm{h}_i, \bm{h}_k) \cdot \bm{h}_k,
\end{equation}

\noindent where the function $\operatorname{att}(\cdot,\cdot)$ evaluates the correlation weight between the current moving state $\bm{h}_i$ and the historical moving state $\bm{h}_k$ through a dot-production method.

% \begin{equation}
% \label{attention_weight}
% \begin{split}
% \small
% \operatorname{att}(\bm{h}_i, \bm{h}_k) &= \frac{(\bm{W}_1 \cdot \bm{h}_k)^T \cdot (\bm{W}_2 \cdot \bm{h}_i)}{\sum_{k^{\prime}=1}^i \exp((\bm{W}_1 \cdot \bm{h}_{k^{\prime}})^T \cdot (\bm{W}_2 \cdot \bm{h}_i))},
% % \alpha_{i, k}& = (\bm{W}_1 \cdot \bm{h}_k)^T \cdot (\bm{W}_2 \cdot \bm{h}_i),
% \end{split}
% \end{equation}
% \noindent where the $\bm{W}_1$ and $\bm{W}_2$ are the parameter matrices to be learned.

Finally, after capturing the enhanced moving state, we construct a topologically constrained linear layer to predict the observed conditional probability $P(l_j | x_{1:i})$. In the topologically constrained linear layer, we introduce a topological adjacency weight $\operatorname{adj}_{l_i, l_j}$ into the linear layer, which is 1 when the candidate road $l_j$ is adjacent to the current road $l_i$ and 0 otherwise. With the topological adjacency weight, we set the conditional probability $P(l_j | x_{1:i})$ of the non-adjacent candidate road $l_j$ to zero, which ensures the continuity of the generated trajectory. The observed conditional probability is formulated as
% As our goal is to generate continuous trajectories, the observed conditional probability $P(l_j | x_{1:i})$ should be limited to zero, where the candidate road $l_j$ is not adjacent to current road $l_i$. To achieve this restriction, we introduce an adjacency weight $\operatorname{adj}_{l_i, l_j}$ into a linear layer, which is 1 when the candidate road $l_j$ is adjacent to the current road $l_i$ and 0 otherwise. Thus, the observed conditional probability is formulated as
% The observed conditional probability is calculated according to Eq.~\ref{observed_probability}.
% Since our goal is to generate fine-grained trajectories, we need to restrict the candidate road segments selected at each step to be contiguous. Inspired by previous work \cite{10.1145/3447548.3467406}, we insert $\operatorname{adj}$ weight into the linear layer to ensure the probability of non-adjacent road segments is zero. The observed probability $g(l_j)$ of a candidate road $l_j$ is calculated by Eq. \ref{g_function}, where the parameter vector $\mathbf{w}_{l_j}$ is the $l_j$-th row of the linear layer parameter matrix $\mathbf{W}_3$.

\begin{equation}
\label{observed_probability}
\begin{split}
\small
%  P(x_{i+1}=l_j \mid x_{1: i}) g(l_j)
 P(l_j \mid x_{1: i})&=\frac{\exp (\bm{w}_{l_j} \cdot \bm{\widetilde{h}}_i \cdot \operatorname{adj}_{l_i, l_j})}{\sum_{l_k} \exp (\bm{w}_{l_k} \cdot \bm{\widetilde{h}}_i \cdot \operatorname{adj}_{l_i, l_k})}, \\
% \text { where } \operatorname{adj}_{x_i, l_j}&= \begin{cases}1, & x_i \text{ and } l_j \text{ is adjacent }\\
% 0, & \text { otherwise }\end{cases}
\end{split}
\end{equation}

\noindent where the parameter vector $\bm{w}_{l_j}$ is the $l_j$-th row of the weight matrix of the linear layer.

The observed cost $g(l_j)$ can be calculated as the negative log of the conditional probability as
% The final observed cost $g(l_j)$ can be calculated according to the chain rule in probability, as shown in Eq.~\ref{function_g} where $l_j$ is the next location of $x_{i+1}$.
% \begin{equation}
% \label{function_g}
% g(l_j) = -\sum_{k=1}^{j} \log P(l_k \mid x_{1: k-1})
% \end{equation}
% \noindent where $l_j$ is the next

\begin{equation}
\label{function_g}
\small
g(l_j) = -\log P(l_j \mid x_{1: i})
\end{equation}

% \begin{equation}
% \begin{aligned}
% P\left(l_{i+1}=l_{j} \mid l_{1: i}\right)=P\left(l_{i+1}\right.&\left.=l_{j} \mid h_{i}\right)=\frac{\exp \left(W \cdot h_{i} \cdot \operatorname{adj}_{l_{i}, l_{j}}\right)}{\sum_{l_{j} \in \mathcal{N}\left(l_{i}\right)} \exp \left(W \cdot h_{i}\right)} \\
% \text { where } a d j_{l_{i}, l_{j}}=& \begin{cases}1, & l_{j} \in \mathcal{N}\left(l_{i}\right) \\
% 0, & \text { otherwise }\end{cases}
% \end{aligned}
% \end{equation}

\subsubsection{Function \textit{h}($\cdot$).} 
To model the expected cost, we assume that a person usually needs to know the relative position of two roads in the road network and the distance between them when estimating the expect cost of getting from a candidate road to the destination. From this perspective, we use a graph attention network to extract the relative position information from the road network and calculate the spherical distance between two road segments. Based on above information, we finally use a multilayer perceptron network to estimate the expected cost $h$.
% Since there is no explicit trajectory information to estimate the expected cost $h(l_j)$, we construct a human movement hypothesis, where a person usually needs to know the location of two roads in the road network and the distance information between them when estimating the expect cost of getting from a candidate road to the destination. From this perspective, we use a graph attention network to extract the road representation from the road network and calculate the spherical distance between two road segments, and finally use a multilayer perceptron Network to estimate the expected cost $h(l_j)$.
% It is more difficult to learn the estimated probability from the candidate road to destination road, since there is no explicit trajectory information. When a person estimates the probability of getting from a candidate road to the destination, he usually needs to know the location of the two roads in the road network and the distance information between them. 

% We build a value network based on GAT, and combine the pre-training mechanism to make the network learn the estimated probability from historical trajectory data. 

% \begin{equation}
%     \mathbf{v_{l_i}} = 
% \end{equation}

First, we build a graph attention network~\cite{GAT} to learn the structural road representation containing the relative position information. The update of the graph attention network can be given as

\begin{equation}
\small
    \bm{N}^{(z+1)} = \operatorname{GAT}(\bm{N}^z, \bm{A}),
\end{equation}
where $\bm{N}^z \in \mathbb{R}^{|L| \times d_s}$ is the matrix of road representation at the $z$-th iteration, the $l_i$-th row $\bm{n}_{l_i} \in \mathbb{R}^{d_s}$ corresponds to the representation of road segment $l_i \in L$, and $\bm{A}$ is the road adjacency matrix.

For initialization, we construct the road contextual embedding vectors $\bm{v}_{l_j}$ and set $\bm{n}^0_{l_i} = \bm{v}_{l_i}$. Given a road segment $l_i$, we consider 6 kinds of road attributes, namely segment length, segment width, max speed limitation, lane number, road type, and longitude and latitude. We perform normalization on continuous attributes and one-hot encoding on discrete attributes, and concatenate the encoded attributes as the contextual embedding vector $\bm{v}_{l_i}$.

% Based on the road attribute matrix $\mathbf{V}$ and the road adjacency matrix $A$ extracted above, we apply the graph attention network~\cite{GAT} to learn the latent road representation $\mathbf{N}_S$.
% \begin{equation}
% \label{gat_representation}
% \mathbf{N_S} = \operatorname{GAT}(\mathbf{V}, \mathbf{A}),
% \end{equation}

After obtaining the road representation, we use a multilayer perceptron network to predict the expected conditional probability from a candidate road $l_j$ to the destination $l_d$ as
% as shown in Eq. \ref{h_probability}.

\begin{equation}
\label{h_probability}
\small
P(l_j \mid l_d) = \operatorname{MLP}(\bm{n}_{l_j}, \bm{n}_{l_d}, d_{l_j, l_d}).
\end{equation}

\noindent where the $\bm{n}_{l_j}$ and $\bm{n}_{l_d}$ is the road representation of road $l_j$ and $l_d$, and the $d_{l_j, l_d}$ is the spherical distance between them.

In the end, the expected cost $h(l_j)$ is calculated as the negative log of the conditional probability $P(l_j | l_d)$ as

\begin{equation}
\label{function_h}
\small
    h(l_j) = -\log P(l_j \mid l_d).
\end{equation}

\subsubsection{Remark.} Compared with the original A* algorithm, our function $f$ is capable to model the diverse human movement policy with the introduction of temporal information and the sequential moving state. Besides, combining the structural road representation, the effectiveness of the expected cost $h(l_j)$ is improved.

% \subsubsection{Final Probability.} In the end, the conditional probability $f(l_j) = P(l_j \mid x_{1:i}, x_d)$ is defined as Eq. \ref{final_probability}, which is weighted sum of observed probability $g(l_j)$ and estimated probability $h(l_j)$. The $w_3$ and $w_4$ are parameters to learn.

% \begin{equation}
% \label{final_probability}
% f(l_j) = w_3 \cdot g(l_j) + w_4 \cdot h(l_j)
% \end{equation}

% introduce the overview of discriminator

\subsection{Enhancing Generator with Discriminator}

In general, the generative adversarial network learned by a min-max game as follows

\begin{equation}
\small
    \min_{\theta} \max_{\phi} \mathbb{E}_{\bm{x}\sim p_d}[\log D_{\phi}(\bm{x})] + \mathbb{E}_{\bm{x}\sim G_{\theta}}[\log (1-D_\phi(\bm{x}))],
\end{equation}
\noindent where $\bm{x}\sim p_d$ denotes the sample from the real data distribution, $\bm{x} \sim G_{\theta}$ denotes the sample generated by the generator, and $D_\phi$ is the $\phi$-parameterized discriminator.

In the min-max game, the goal of the discriminator $D_\phi$ is to distinguish the input sample, and the generator $G_\theta$ optimizes itself based on the discriminator's output signal, i.e., the reward $R$ as

\begin{equation}
\small
D_\phi(\bm{x}) = R(\bm{x}).
\end{equation}

In our framework, the input sample $\bm{x}$ is a trajectory $T$, which is a time-series with spatial information. Thus, we distinguish the input trajectory in terms of the time-series similarity and the spatial similarity, and generate corresponding reward $R$ as

\begin{equation}
\small
    R(T) = R_s(T) + R_m(T),
\end{equation}

\noindent where $R_s$ denotes the sequential reward from the time-series aspect, where we evaluate the similarity based on the hidden sequential transition pattern extracted by an LSTM network. $R_m$ denotes the mobility yaw reward from the spatial aspect, where we evaluate the similarity according to the trajectory's mobility yaw distance from the real trajectories.

\subsubsection{Sequential Reward.} To extract the sequential transition pattern, we build a sequential discriminator $D_s$ following the same steps in the function $g$: we first embed the input trajectory, and then obtain the hidden sequential transition pattern $\bm{h}$ based on an LSTM network. Then, we use a linear layer to predict the probability that the trajectory is real, which is the sequential reward $R_s$ as

\begin{equation}
\small
\label{classification_reward}
\begin{split}
R_s(T)= \begin{cases} D_{s}(T) & \text { if } T \text{ is finished;} \\
\frac{1}{N} \sum_{n=1}^{N} D_{s}(\mathrm{MC}(T)) & \text{ else,}
\end{cases}
\end{split}
\end{equation}

\noindent where we further leverage the Monte Carlo search to evaluate the average sequential reward for the intermediate step in the trajectory.

This is because the trajectory from the generator is sequentially generated, which requires the discriminator to provide rewards for each step of the trajectory. However, current discriminator can not provide proper reward for the intermediate step. For the intermediate step $i$, the current trajectory $x_{1:i}$ is still unfinished. This means that the reward, based on current sub-trajectory $x_{1:i}$, only considers the sequential transition pattern of the current sub-trajectory but ignores the future outcome. Thus, following the previous work~\cite{SeqGAN}, we apply Monte Carlo search to evaluate the intermediate step. To do this, we first maintain an older version generator $G_{old}$, which is the last step version of the current generator. Then, we use the $G_{old}$ to complete the current sub-trajectory by repeating Monte Carlo search $N$ times. The complete trajectories are then fed into the $D_s$ to generate sequential reward $R_s$ for the intermediate step.

\subsubsection{Mobility Yaw Reward.} The mobility yaw distance of a input trajectory $T$ is defined as the minimum distance to the real trajectories set $S_{od}$ in the form of

\begin{equation}
\small
\label{yaw_distance}
dis = \min\{\operatorname{DTW}(T, T^\prime) \mid T^\prime \in S_{od}\},
\end{equation}

\noindent where we leverage the widely-used trajectory distance metric DTW~\cite{dtw} to measure the mobility yaw distance, and the real trajectories set $S_{od}$ contains the  real trajectories shared the same OD with the input trajectory $T$.

Furthermore, we calculate the mobility yaw distance for each step of the trajectory $T$ by the Monte Carlo search. The mobility yaw reward is defined as the change in yaw distance $\Delta dis_i = dis_i - dis_{i-1}$ as

\begin{equation}
\small
\label{yaw_reward}
R_m(T) = \operatorname{norm}(\Delta dis_i),
\end{equation}
\noindent where we further apply a normalization to dimensionless the change of mobility yaw distance.

The change in yaw distance $\Delta dis_i$ indicates how much the current step's mobility yaw distance has decreased compared to the previous step. The more the mobility yaw distance is decreased, the more correct the decision of the current step is. Correspondingly, the reward for the current step should be larger.

\subsection{Two-Stage Generation Process}

% There are two problems for generating continuous trajectory with stochastic generation process. First, the space complexity of generating continuous trajectory on the road network is greater than that of generating ordinary trajectory.

% In the traditional stochastic generation process, the next road segment in trajectory is randomly chosen according to the conditional probability $\pi(a | s)$ predicted by the generator. However, due to the sparsity of trajectory data and the complexity of road network, it is hard for the generator to give correct predictions at every intersection. Thus, once the generator predicts wrong, the stochastic generation process will continue to generate under the wrong premise state, which will make it difficult to reach the stationary state (the destination). Especially, when generating long trajectories, the probability of the generator making an error increases as the number of predictions made by the generator increases.

The stochastic generation process has the problem of error accumulation, where trajectory is randomly generated according to the probability given by the generator as

\begin{equation}
\label{stochastic_generation}
\small
    P_{\theta}(\hat{T})=\prod_{i=1}^{n} \pi(a_i | s_i) = \prod_{i=1}^{n} P_{\theta}(l_{i+1} | x_{1:i}, l_d).
\end{equation}

However, once the generator predicts wrong, the stochastic generation process continues to generate under the wrong premise state. Especially, when generating long trajectories, the probability of the generator making an error increases as the number of predictions made by the generator increases. This makes it difficult to reach the destination and reduces the quality of the generated trajectory.
%  and the errors accumulate

To solve the above problem, we propose a novel two-stage generation process based on A* search. We first construct the structural regions above the road network. Then, we generate the regional trajectory in the first stage. In the second stage, we generate the continuous trajectory in the guidance of the regional trajectory. With the A* search, our generation process can roll back to correct the error, when the generator predicts incorrectly. Because, in the A* search, all possible trajectories currently searched are maintained, and the trajectory with the least cost (highest probability) is selected for searching each time. In addition, with the two-stage generation, not only the space and time complexity of the generation process is reduced, but the generation of long trajectories is also avoided.

\subsubsection{Building Structural Region.} We leverage the multilevel graph partitioning algorithm KaFFPa~\cite{KaHIP} to construct structural regions on the top of the road network. It partitions the graph into $k$ blocks under the constraint that the number of edges between blocks is minimized and the maximum block size does not exceed $1+\epsilon$ times the average block size. With the KaFFPa algorithm, we can obtain relatively independent structural regions of similar size, and the mapping relationship between road and region. Formally, we describe the mapping relationship between road and region as the mapping matrix $\bm{M} \in \mathbb{R}^{|L| \times |R|}$, where each element is defined as

\begin{equation}
\small
\label{mapping_matrix}
m_{l_i,r_j}= \begin{cases}1, & l_i \in r_j; \\ 0, & \text { other,}\end{cases}
\end{equation}

\noindent where the $l_j$ is the $i$-th road segment in road segment set $L$, and the $r_j$ is the $j$-th structural region in region set $R$.

% , and then obtain the road to region mapping matrix $\mathbf{M}$, where each element is defined as
%  Eq. \ref{mapping_matrix}.

% As mentioned above, the human mobility behavior at the region level is modeled by the same generative adversarial network structure as the human mobility behavior at the road level. However, it is worth noting that the region contextual embedding matrix $\mathbf{V}_R$ is calculated from the road contextual embedding matrix $\mathbf{V}$. In practice, 

Based on the mapping matrix $\mathbf{M}$, the region-level data can be obtained through mapping from the road-level data. Then, we use the same proposed adversarial generative framework to learn the region-level human movement policy. See the appendix for more details.
% For example, the road-level trajectory $T$ can be mapped to a region-level trajectory $T_R$ by mapping each road $l_i$ in $T$ to corresponding region $r_i$. The region attributes matrix $\mathbf{V}_R$ is also mapping from the road attribute matrix as
% , as shown in Eq.~\ref{region_matrix}. 
% \begin{equation}
% \label{region_matrix}
% \mathbf{V}_R = \mathbf{M}^T \cdot \mathbf{V}.
% \end{equation}

% The only difference between the two is in the network input.

\subsubsection{Regional Trajectory Generation.} In the first stage, we aim to generate the regional trajectory. At the beginning, we sample a tuple of origin and destination $(l_s, l_d)$ and the start timestamp $t_s$ from the historical OD matrix, which can be counted from the given real-world trajectory dataset. Then, we map the road-level origin and destination to the region-level origin and destination $(r_s, r_d)$ according to the road-to-region mapping relationship, where we map the $l_s$ to $r_s$ only if the $m_{l_s, r_s} = 1$. With the region-level origin and destination and the start timestamp as input, we generate regional trajectory based on the A* search and the region-level generator.

\subsubsection{Road-level Continuous Trajectory Generation.} In the second stage, we generate the continuous trajectory in the guidance of the regional trajectory. In practice, we construct the boundary road segments set $S_{r_i, r_j}$ between each region $r_i$ and region $r_j$, and count the visiting frequency of each boundary road segment in advance. Then, the regional trajectory can be mapped to the road-level, according to the Algorithm \ref{map_regional_trajectory}.

\begin{algorithm}[tb]
\caption{Process of Mapping the Regional Trajectory}
\label{map_regional_trajectory}
\begin{algorithmic}[1] %[1] enables line number
\REQUIRE The regional trajectory $T_R = \{r_1, r_2, \ldots, r_n\}$ and the origin $l_s$ and destination $l_d$.
\ENSURE The discontinuous road segment trajectory $T$.
\STATE Initialize the $T$ as $\{l_s\}$.
\FOR{ $i = 1$ to $n-1$}
    \STATE Randomly choose a boundary road segment $l$ from the boundary road segments set $S_{r_i, r_{i+1}}$ based on the historical visit frequency.
    \STATE Append the boundary road segment $l$ to the $T$.
\ENDFOR
\STATE Append the $l_d$ to the $T$.
\STATE Return $T$.
\end{algorithmic}
\end{algorithm}

After obtaining the discontinuous road segment trajectory $T$, we use the road-level generator to complete the trajectory between each pair $(l_i, l_{i+1})$ in the $T$, which is the final continuous trajectory. In this way, we only need to generate in a small structural region instead of the huge road network.

% \subsubsection{Generation Process.} As presented in Figure \ref{two_stage_generation}, our two-stage generation process consists of four steps: (1) First, sample a tuple of origin and destination $(x_s, l_d)$ from the historical OD matrix, and then map the road-level OD to the region-level OD according to the road-to-region mapping relationship. (2) Second, generate the region-level trajectory using the A* search and the region generator. (3) Third, map the region-level trajectory to coarse-grained trajectory. In detail, we count the boundary road segments between each region in advance. The region-level trajectory can be mapped as coarse-grained trajectory composed of corresponding boundary road segments. In particular, for the case where there are multiple boundary road segments between two regions, we randomly select the mapped road according to the historical visit frequency. (4) 

\subsection{Training Process}
% Considering the data noise and complicated characteristics of mobility trajectory, it is difficult to obtain the promising performance when training the whole framework from scratch. To improve the efficiency and effectiveness of the training process, we first pre-train our framework with tasks related to human mobility behavior. In practice, We pre-train \textit{function g} and \textit{function h} with trajectory next-location prediction task separately, as they model human movement policy from different perspectives. For the discriminator, we directly use the trajectory binary classification task for pre-training.

% In this way, we can take full advantage of the mobility data and enable the framework to preview important human mobility regularities before the GAN training.
At the beginning of the training, we use the trajectory next-location prediction task to pre-train the function $g$ and $h$ separately, as they model human movement policy from different perspectives. Then, we pre-train the sequential discriminator based on the generated trajectories and the real trajectories. After the pre-training, we follow the REINFORCE algorithm~\cite{williams1992simple} to generate the policy gradient by receiving the reward $R(T)$ from the discriminator $D$ as
% , as shown in Eq. \ref{policy_gradient}.
% from the generator $G_\theta$, from the training set

\begin{equation}
\label{policy_gradient}
\small
\begin{aligned}
\nabla_{\theta}=\nabla_{\theta} \mathbb{E}_{p_{\theta}(T)}[R(T)]=\mathbb{E}_{p_{\theta}(T)}[R(T) \nabla_{\theta} \log p_{\theta}(T)],
\end{aligned}
\end{equation}

\noindent where $\theta$ is the parameter of the generator $G$, $T$ is the generated trajectory, the reward $R(T)$ is the sum of sequential reward $R_s$ and mobility yaw reward $R_m$ from the discriminator $D$. Based on the policy gradient $\nabla_{\theta}$, parameter $\theta$ is updated by $\theta \leftarrow \theta+\alpha \nabla_{\theta}$, where $\alpha$ is the learning rate.

% \subsubsection{Reinforcement Learning.}

% \subsubsection{Pre-training Mechanism.}

%% file: experiment.tex
\section{Experiment}

We first evaluate \name in terms of macro-similarity and micro-similarity between the synthetic trajectories and the real trajectories. Next, we conduct two case studies: the trajectory next-location prediction and the traffic control simulation, to assess the utility of synthetic trajectories.
% other than the analysis of statistical similarity

\subsection{Experiment Setting}

\subsubsection{Dataset.} We use two real-world mobility dataset to measure the performance of our proposed framework. The \textit{BJ-Taxi} dataset contains real GPS trajectory data of Beijing taxis for one week from November 1 to 7, 2015 within Beijing's Fourth Ring Roads, which is sampled every minute. The \textit{Porto-Taxi} dataset is originally released for a Kaggle trajectory prediction competition with a sample period of 15 seconds. For the two datasets, we collect corresponding road network information from the open street map, and then perform map maching~\cite{fmm} to obtain real continuous trajectories. To eliminate abnormal trajectories, we remove trajectories with lengths less than 5 and trajectories with loops. The statistics of the two datasets after preprocessing are shown in Table~\ref{dataset}. For each trajectory dataset, we randomly split the whole dataset into three parts in the ratio of 6:2:2: a training set, a validation set, and a test set.

% by aligning GPS points with road segments in the road network

% To measure the performance of our proposed model, We collected real GPS trajectory data for one week from November 1 to 7, 2015 within Beijing's Fourth Ring Road through Beijing taxis. Then we collect corresponding road network information from open street map, and further perform map maching\cite{fmm} by aligning GPS points with road segments in the road network. To eliminate abnormal trajectories, we remove trajectories with lengths less than 5 and trajectories with loops. The statistics of the \textit{Beijing Taxi} dataset after preprocessing are shown in Table \ref{tab:dataset}.

\begin{table}[tbp]
  \centering
  \small
  \begin{tabular}{ccc}
    \hline
    Data Statistics  & BJ-Taxi & Porto-Taxi\\
    \hline
    Number of Taxis & 15,642 &  435\\
    Number of Roads & 40,306 &  11,095\\
    Number of Intersections & 16,927 & 5,184\\
    Number of Trajectories & 956,070 & 695,085\\
    \hline
  \end{tabular}
  \caption{Basic statistics of two real-world datasets.}
  \label{dataset}
\end{table}

% For each trajectory dataset, we randomly split the whole dataset into three parts in the ratio of 6:2:2: a training set for training models, a validation set for finding the best parameters of models, and a testing set for the final evaluation on various metrics.

\subsubsection{Comparative Benchmarks.} We compare \name with five baselines to validate the performance of our proposed framework: SeqGAN~\cite{SeqGAN}, SVAE~\cite{SVAE}, MoveSim~\cite{MoveSim}, TSG~\cite{TSG}, and  TrajGen~\cite{trajgen}. For more information about the baselines, please refer to the appendix.

% \begin{itemize}
% \item SeqGAN~\cite{SeqGAN}: This method is the classical sequence generation method, which combines policy gradient with GAN to solve sequence generation problem.
% % to solve discrete sequence generation problem We directly apply this method to generate the road location sequence.
% \item SVAE~\cite{SVAE}: This method is the first to combine the variational autoencoder with the Seq2Seq model to generate mobility trajectory.
% \item MoveSim~\cite{MoveSim}: This method builds a self-attention-based generator and designs a mobility regularity-aware loss for the discriminator.
% \item TSG~\cite{TSG}: This method first uses a CNN-based GAN to generate grid-based trajectories, and then uses a BiLSTM-based GAN to generate final trajectories.
% % This method proposes a two-stage trajectory generation GAN. First, it uses a CNN-based GAN to generate grid-based trajectories. Then, it uses a BiLSTM-based GAN to generate final trajectories.
% \item TrajGen~\cite{trajgen}: This method maps trajectories to images, and then uses DCGAN to generate images. Finally, it extracts locations from the image and uses a Seq2Seq model to infer the real trajectory.
% \end{itemize}

\subsubsection{Performance Criteria.} From the macro perspective, we mainly consider the overall distribution of the trajectory dataset. Following the common practice in previous works~\cite{trajgen, MoveSim}, we quantitatively assess the quality of the generated data by computing the similarity of 4 important mobility patterns between the generated and real data: the travel distance~\cite{distance}, the radius of gyration~\cite{radius}, the location distribution~\cite{location_distribution} and the OD flow~\cite{od_flow}. To obtain quantitative results, we use Jensen-Shannon Divergence to calculate the similarity on the four metrics. Refer to the appendix for more information on the metrics.

% 
% \begin{itemize}
%     \item Distance: travel distance~\cite{distance}, which represents the spatial length of a trajectory.  
%     \item Radius: radius of gyration~\cite{radius}, which represents the spatial travel range of a trajectory.
%     \item Location Distribution: visit frequency distribution of each single road~\cite{location_distribution}, which indicates the popularity of roads.
%     \item OD Flow: origin-destination flow~\cite{od_flow}, which represents the overall travel demand of the city. 
% \end{itemize}

From the microscopic perspective, we focus on measuring the similarity between the real trajectories and the generated trajectories with the same OD. In practice, we use three widely used trajectory distance metrics to measure the micro similarity: Hausdorff distance~\cite{hausdorff}, DTW~\cite{dtw}, and EDR~\cite{edr}. We randomly select 5,000 trajectories from the test set, and make the generative model generate the corresponding trajectory. Finally, the average distance between the generated trajectories and the real trajectories is taken as the evaluation result.
% We first randomly select 5000 trajectories from the test set. Then, the origin and destination of each real trajectory are used as input to make the generative model generate the corresponding trajectory. Finally, the average distance between the generated trajectories and the real trajectories is taken as the evaluation result. It is worth noting that some baseline models (TrajGen, TSG, and SeqGAN) can not take an OD as input. In this case, we directly generate trajectories from the baseline models.

\subsection{Evaluation Results}

The evaluation results of macro-similarity and micro-similarity are shown in Table~\ref{similarity_result}. In each row, the best result is highlighted in boldface and the second best is underlined. From the statistics, we can first conclude that our proposed \name significantly outperforms all baselines in terms of both macro-similarity metrics and micro-similarity metrics.

\begin{table*}[tbp]
  \centering
    \resizebox{0.97\linewidth}{!}{
        \begin{tabular}{lcccc|ccc|cccc|ccc}
        \toprule
        \multicolumn{1}{c}{\multirow{3}[6]{*}{Model\textbackslash Metrics}} & \multicolumn{7}{c|}{BJ-Taxi}                          & \multicolumn{7}{c}{Porto-Taxi} \\
        \cmidrule{2-15}          & \multicolumn{4}{c|}{Macro-Similarity Metrics} & \multicolumn{3}{c|}{Micro-Similarity Metrics} & \multicolumn{4}{c|}{Macro-Similarity Metrics} & \multicolumn{3}{c}{Micro-Similarity Metrics} \\
        \cmidrule{2-15}          & \multicolumn{1}{c}{Distance} & \multicolumn{1}{c}{Radius} & \multicolumn{1}{c}{Location Frequency} & \multicolumn{1}{c|}{OD Flow} & \multicolumn{1}{c}{Hausdorff} & \multicolumn{1}{c}{DTW} & \multicolumn{1}{c|}{EDR} & \multicolumn{1}{c}{Distance} & \multicolumn{1}{c}{Radius} & \multicolumn{1}{c}{Location Frequency} & \multicolumn{1}{c|}{OD Flow} & \multicolumn{1}{c}{Hausdorff} & \multicolumn{1}{c}{DTW} & \multicolumn{1}{c}{EDR} \\
        \midrule
        SeqGAN & 0.4945  & 0.4717  & \underline{0.1124}  & 0.6519  &    13.85   &  \underline{199.51}  & 0.9986  & 0.3504   &   0.3030    &  0.1426     & 0.3370  & 4.19  &  \underline{82.55} & 0.9950 \\
        SVAE  & 0.3027  & 0.2356  & 0.1531  & 0.6825  & \underline{10.88}  & 430.88  & 0.9977  &    0.3608  &   0.1866   &  \underline{0.1342}  &   0.6274   &  \underline{3.36}   & 133.15 & 0.9794 \\
        MoveSim & 0.5370  & 0.4073  & 0.4062  & 0.6362  & 13.41  & 240.16  & \underline{0.9452}  &    0.4732   &  0.2020 &  0.3158  &  \underline{0.2496}   &  4.17 &   101.18   & \underline{0.9531}  \\
        TSG   & 0.6084  & \underline{0.0952}  & 0.5362  & 0.6507  &  11.07  &   2446.58  &   0.9991  & 0.5088     &  0.1278    &    0.4165   &    0.5527   & 3.88 &704.39  &0.9967 \\
       TrajGen & \underline{0.2248}  & 0.1376 & 0.2553  & \underline{0.6321}  &   11.65   & 427.75  &   0.9970  & \underline{0.1362} & \underline{0.0653}  & 0.3652 &  0.5519 & 4.37 & 198.90 & 0.9934\\
        Our   & \textbf{0.0054}  & \textbf{0.0015}  & \textbf{0.0763}  & \textbf{0.0000}  & \textbf{0.76}  & \textbf{13.71}  & \textbf{0.4522}  &  \textbf{0.0081}    &  \textbf{0.0021}    & \textbf{0.1203}  &  \textbf{0.0000}  &  \textbf{0.65}  & \textbf{18.12}  & \textbf{0.5580} \\
        \bottomrule
      \end{tabular}%
    }
  \caption{Performance comparison of our model and baselines on two real-world datasets, where lower results are better.}
  \label{similarity_result}%
\end{table*}%

\begin{figure}[tbp]
  \begin{subfigure}{0.45\columnwidth}
    \centering
    \includegraphics[width=\columnwidth]{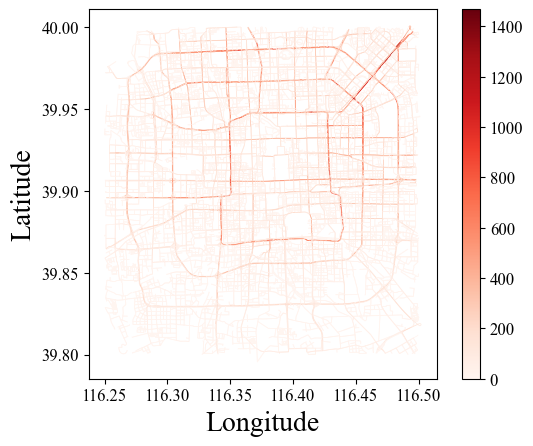}
    \caption{Real Data}
    \label{true heat map}
  \end{subfigure}\hfill
  \begin{subfigure}{0.45\columnwidth}
    \centering
    \includegraphics[width=\columnwidth]{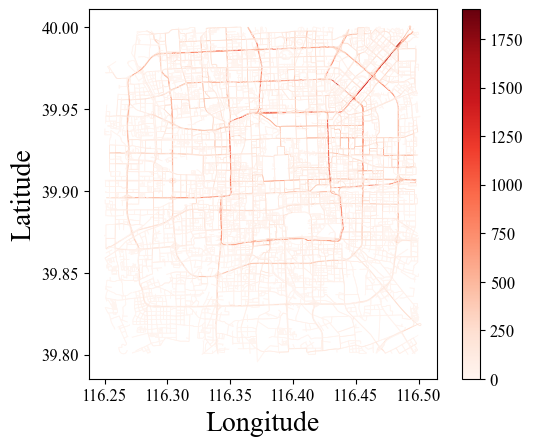}
    \caption{Our Data}
    \label{our heat map}
  \end{subfigure}\par
%   \begin{subfigure}{0.47\columnwidth}
%     \centering
%     \includegraphics[width=\columnwidth]{figure/seqgan_road_heatmap.png}
%     \caption{SeqGAN Data}
%     \label{seqgan heat map}
%   \end{subfigure}
  \begin{subfigure}{0.45\columnwidth}
    \centering
    \includegraphics[width=\columnwidth]{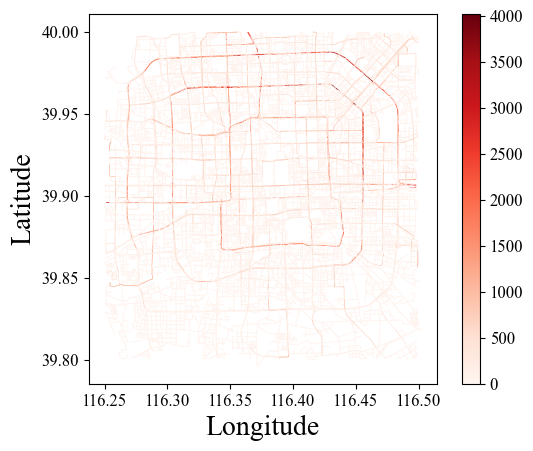}
    \caption{SVAE Data}
    \label{svae heat map}
  \end{subfigure}\hfill
%   \begin{subfigure}{0.47\columnwidth}
%     \centering
%     \includegraphics[width=\columnwidth]{figure/MoveSim_road_heatmap.png}
%     \caption{MoveSim Data}
%     \label{movesim heat map}
%   \end{subfigure}
%   \begin{subfigure}{0.47\columnwidth}
%     \centering
%     \includegraphics[width=\columnwidth]{figure/tsg_road_heatmap.png}
%     \caption{TSG Data}
%     \label{tsg heat map}
%   \end{subfigure}
  \begin{subfigure}{0.45\columnwidth}
    \centering
    \includegraphics[width=\columnwidth]{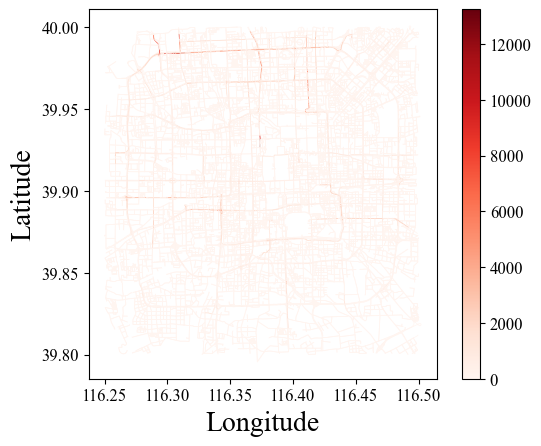}
    \caption{TrajGen Data}
    \label{trajgen heat map}
  \end{subfigure}
  \caption{Visualized \textit{location distribution} of different trajectory data.}
  % The darker the road, the more frequently it is visited.
  \label{heat map}
\end{figure}

From the perspective of macro-similarity, compared with the second best model, \name improves the \textit{Distance}, \textit{Radius}, and \textit{Location Frequency} metrics by 97.57\%, 98.94\% and 70.09\% respectively in \textit{BJ-Taxi}. As for the \textit{OD Flow} metric, since we take both the origin and destination as input, \name achieve to completely restore the real OD flow state of the city. Among the baselines, TrajGen performs best because the trajectories it generates are derived from images, which enables it to generate continuous trajectories as the image resolution increases. While other sequential generative baselines fail to model the continuous mobility pattern. 
% , which is impossible for other baseline models.  to a certain extent
Besides, to better illustrate the detailed difference, we leverage heat map to visualize the \textit{location distribution} of the top three generated data in Beijing, as shown in Figure~\ref{heat map}.
% Specifically, we render the road into different color according to its visit probability, that is, the darker the road color, the higher the visit frequency.

%  First, we leverage heat map to visualize the location distribution of the trajectory dataset from a qualitative perspective. Specifically, we count the visit frequency of each road in the road network, and then render the road into different color according to its visit frequency. That is, the darker the road color, the higher the visit frequency. 

As for the micro-similarity, our proposed \name achives to simulate the individual trajectory, and shows clear advantages over others, which is mainly because other baseline models ignore the impact of destination on the human mobility behavior.

\subsection{Case Study}

% \subsubsection{Case Study 1: Human Mobility Law Verification.}

\subsubsection{Case Study 1: Trajectory Next-Location Prediction.} Trajectory next-location prediction task aims to predict the location of mobility trajectory through mining the human mobility patterns. Therefore, we can leverage this task to examine whether there are real mobility patterns in the generated data based on \textit{BJ-Taxi} dataset.

In this case study, we choose DeepMove~\cite{DeepMove} as the trajectory next-location prediction model. We mix the generated data with the real data in various ratios and train the DeepMove model using the mixed data. Then, we test the model performance on real data and calculate the performance difference from the model trained entirely on real data.
%  Thus, the utility of the generated data is verified.  to mine the human mobility patterns

\begin{figure}[t]
%   \hfill
  \begin{subfigure}{0.45\columnwidth}
    \centering
    \includegraphics[width=\columnwidth]{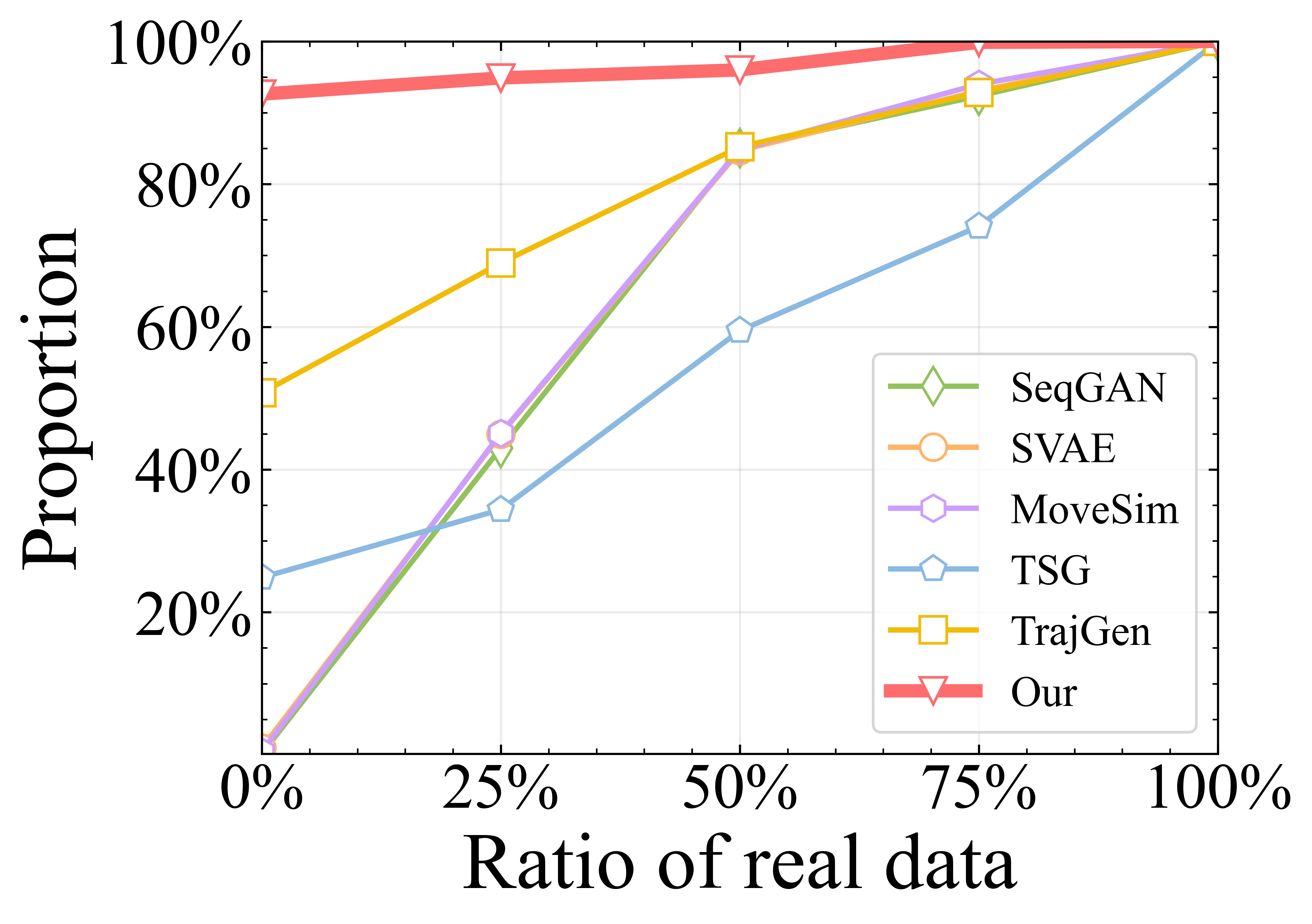}
    \caption{Recall@5}
    \label{recall_5}
  \end{subfigure}\hfill
  \begin{subfigure}{0.45\columnwidth}
    \centering
    \includegraphics[width=\columnwidth]{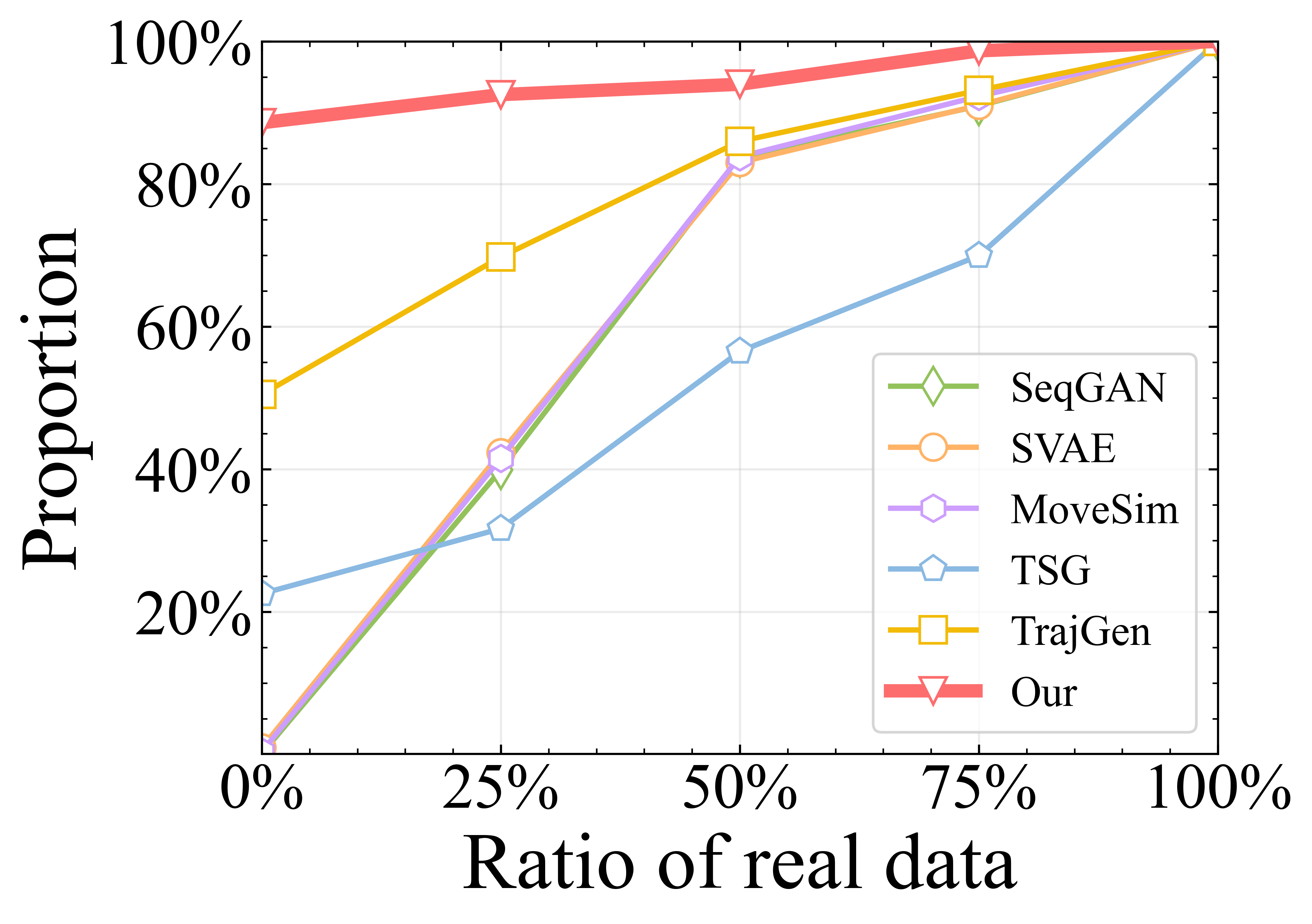}
    \caption{NDCG@5}
    \label{ndcg_5}
  \end{subfigure}
  \caption{The proportion of performance of the model trained on mixed data over the performance of the model trained on real data.}
  \label{trajectory_next_location_figure}
\end{figure}

Figure~\ref{trajectory_next_location_figure} shows the performance difference between the model trained on the mixed data and the model trained on the real data, e.g., only the real data is used for training when the $x$ value is 100\%. We can conclude the utility of our generated data exceeds the data generated by other models. When only trained with our generated data, the DeepMove achieves 92.61\% and 88.59\% performance of the DeepMove trained on real data on the \textit{Recall@5} and \textit{NDCG@5} metrics respectively. However, among the baselines, the best proportions of performance are 50.74\% and 50.47\% respectively.
% , and only the generated data is used for training when the $x$ value is 0\%.

\subsubsection{Case Study 2: Traffic Control Simulation.} Traffic simulation is an important application of trajectory generation. Thus, we choose a case of Beijing Sanyuan Bridge traffic control to verify that our framework is competent for the human mobility simulation task.

\begin{figure}[t]
  \begin{subfigure}{0.32\columnwidth}
    \centering
    \includegraphics[width=\columnwidth]{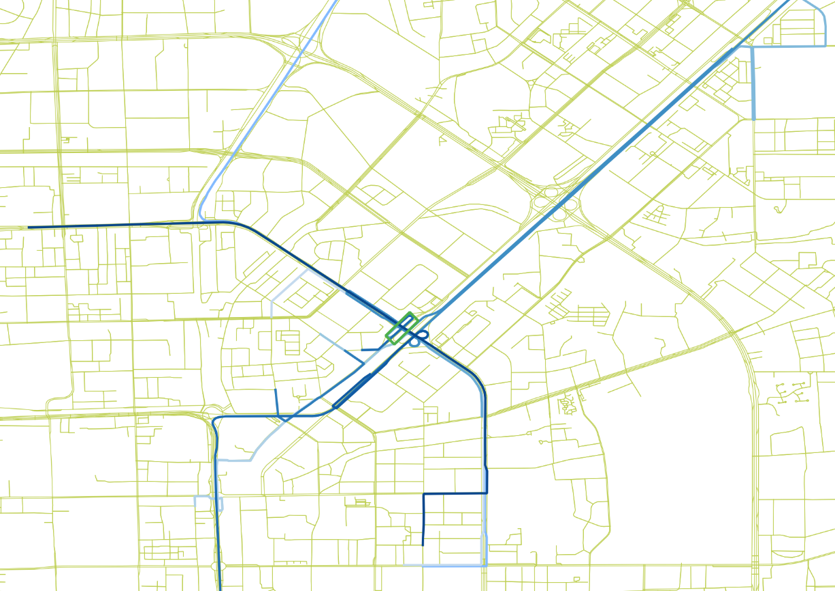}
    \caption{Normal situation}
    \label{without_traffic_control}
  \end{subfigure}\hfill
  \begin{subfigure}{0.32\columnwidth}
    \centering
    \includegraphics[width=\columnwidth]{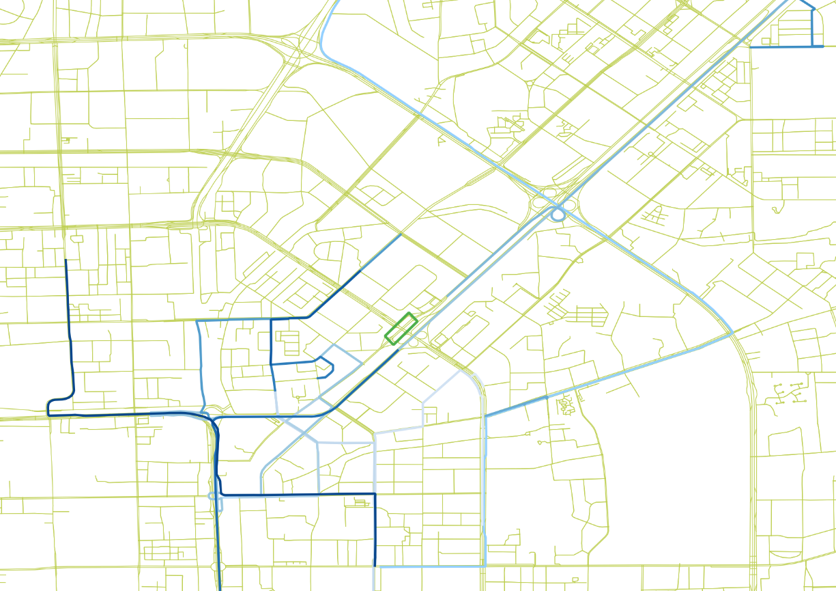}
    \caption{Traffic control}
    \label{traffic_control}
  \end{subfigure}\hfill
  \centering
  \begin{subfigure}{0.32\columnwidth}
    \centering
    \includegraphics[width=\columnwidth]{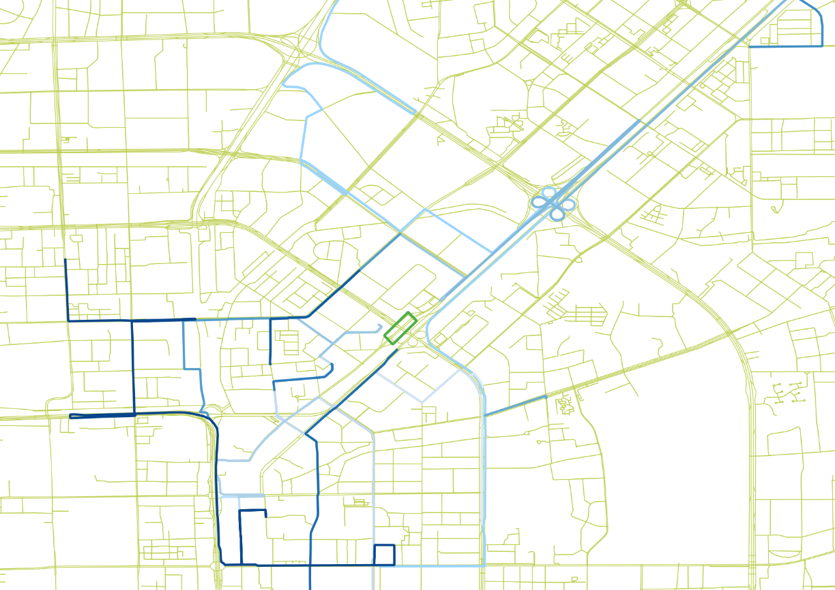}
    \caption{Simulation result}
    \label{simulate_traffic_control}
  \end{subfigure}
  \caption{Visualized vehicle trajectories near Sanyuan Bridge \ref{without_traffic_control}) when there is no traffic control,  \ref{traffic_control}) when there is traffic control, and \ref{simulate_traffic_control}) which is the simulation result.}
  \label{traffic_control_visualize}
\end{figure}

% , prohibiting the passage of motor vehicles except buses
On November 13, 2015, the Sanyuan Bridge on the Third Ring Road in Beijing was imposed a one-day traffic control due to the construction reason. We randomly visualized 15 trajectories without and with traffic control near the Sanyuan Bridge, as shown in Figure~\ref{without_traffic_control} and Figure~\ref{traffic_control}, where the green lines are the Beijing road network, the blue lines are the trajectories of different vehicles, and the area in the green box is the Sanyuan Bridge.

% It is not difficult to conclude from the figures that the traffic control has indeed affected the passage of surrounding vehicles.

In this case study, we set the roads under construction as unreachable and then simulate trajectories nearby Sanyuan Bridge. The part of simulation result is visualized in Figure \ref{simulate_traffic_control}. In addition, we quantitatively calculate the micro similarity between real trajectories and generated trajectories. However, since none of the above baseline models can handle the traffic control situation, we chose two route planning models as baseline models in this case study, i.e., NASR\cite{NASR} and A* search\cite{A_star}. The evaluation result is shown in Table~\ref{simulate_micro_similarity}. From the test result, our \name is closer to the real traffic control situation than existing route planning method NASR, and achieves significant progress on the first two metrics.
% , to simulate the traffic control situation
% In order to quantitatively analyze the performance of the simulation
\begin{table}[htbp]
  \centering
  \small
    \begin{tabular}{cccc}
    \hline
          & Hausdorff & DTW   & EDR \\
    \hline
    A* search  & 2.7599 & 49.3791 & 0.7160 \\
    % \hline
    NASR & \underline{2.0393} & \underline{37.1353} & \underline{0.5658} \\
    % \hline
    Our   & \textbf{0.9790} & \textbf{16.3914} & \textbf{0.5451} \\
    \hline
    \end{tabular}%
  \caption{The micro similarity between real trajectories and simulated trajectories.}
  \label{simulate_micro_similarity}%
\end{table}%

%% file: related_work.tex
\section{Related Work}

The trajectory generation problem takes an origin road segment and an destination road segment as input, and outputs the corresponding trajectory.
Existing works of trajectory generation can be classified into two groups: model-based methods and model-free methods. Based on assumptions about the human mobility regularity, model-based methods model the individual mobility by limited parameters. For example, TimeGeo~\cite{TimeGeo} leverages data-driven methods to estimate the parameters from the real data, and then generate trajectory via Markov based models with the simplified assumption of human mobility. While achieving promising performance in some cases, these methods with simplified assumptions can not model the complex mobility in reality.
% The synthetic trajectory is generated by the agent switching between the two mechanisms described above. Several studies subsequently improved EPR by adding increasingly sophisticated spatial or social mechanisms\cite{}. 
% For example, EPR\cite{epr} assumes that the human mobility is governed by two mechanisms: an exploration mechanism, in which the agent selects a new location that have never been visited based on a random walk process; and a preferential return mechanism, in which the agent returns to previously visited locations based on its frequency.

Recently, the researchers turned to apply model-free methods like VAE and GAN to generate the mobility trajectory. SAVE~\cite{SVAE} is a combination of VAE and Seq2Seq model, which combines the ability of VAE to learn the underlying distribution of real data and the ability of LSTM to process sequential data. MoveSim~\cite{MoveSim} is a model-free GAN framework that integrates the domain knowledge of human mobility regularity. TrajGen~\cite{trajgen} maps a trajectory into an image and utilizes a CNN-based GAN to generate the virtual trajectory image. Then, TrajGen applies a Seq2Seq model to infer the real trajectory sequence from the location sequence extracted from the virtual image. Although these methods do not rely on the simplified human mobility assumptions, but instead directly learned the underlying distribution of the data. However, these methods igonre the problem of the continuity of the generated trajectory, which is crucial in practical human mobility simulation applications. Hence, we propose a novel two-stage generative adversarial framework to solve this problem.
% However, these methods ignore to capture the fine-grained human mobility behavior. This makes them fail to generate fine-grained trajectories, which are required in human mobility simulation applications. Hence, we propose a novel two-stage generative adversarial framework to solve this problem.

%% file: conclusion.tex
\section{Conclusion}

In this paper, we propose a novel two-stage generative adversarial framework to solve the continuous trajectory generation problem, which combines the advantages of human mobility prior knowledge with the model-free learning paradigm. Although many previous works have investigated the trajectory generation problem, we are the first to directly generate continuous trajectories on the road network. The extensive experiments demonstrate that our framework outperforms five state-of-the-art baselines significantly, including aspects of macro-similarity, micro-similarity and data utility. Besides, through a traffic control simulation case study, we prove that our framework can be directly applied to real-world city traffic simulations. As future work, we will further explore other potential factors of human mobility and extend the simulation to various applications.
%  and achieve promising performance